\documentclass[letterpaper]{article} % DO NOT CHANGE THIS
\usepackage{aaai2027}  % DO NOT CHANGE THIS

\nocopyright
\usepackage[hyphens]{url}  % DO NOT CHANGE THIS
\usepackage{graphicx} % DO NOT CHANGE THIS
\def\UrlFont{\rm}  % DO NOT CHANGE THIS
\usepackage{natbib}  % DO NOT CHANGE THIS AND DO NOT ADD ANY OPTIONS TO IT
\usepackage{caption} % DO NOT CHANGE THIS AND DO NOT ADD ANY OPTIONS TO IT
\usepackage{amsmath}
\usepackage{algorithm}
\usepackage{algorithmic}

\usepackage{booktabs}
\usepackage{colortbl}
\usepackage{framed}
\definecolor{tablegroupgray}{RGB}{242,242,242}
\definecolor{tablemethodblue}{RGB}{229,241,251}

\title{AgentPatch: Coarse-to-Fine Weak-Task Repair for Merging Agentic Multimodal Large Language Models}
\author{
Zibo Shao\textsuperscript{\rm 1,\rm 2,\rm 3},
Baochen Xiong\textsuperscript{\rm 1,\rm 2,\rm 3},
Chengdong Xu\textsuperscript{\rm 4},
Linhui Xiao\textsuperscript{\rm 2},\\
Kaichen Li\textsuperscript{\rm 1,\rm 2,\rm 3},
Haoran Gong\textsuperscript{\rm 2},
Yan Li\textsuperscript{\rm 2},
Yaguang Song\textsuperscript{\rm 2}\corresponding,
Xiaoshan Yang\textsuperscript{\rm 1,\rm 2,\rm 3}\corresponding
}
\affiliations{
\textsuperscript{\rm 1}State Key Laboratory of Multimodal Artificial Intelligence Systems, Institute of Automation,\\
Chinese Academy of Sciences, Beijing, China\\
\textsuperscript{\rm 2}Pengcheng Laboratory, Shenzhen, China\\
\textsuperscript{\rm 3}School of Artificial Intelligence, University of Chinese Academy of Sciences, Beijing, China\\
\textsuperscript{\rm 4}Sun Yat-Sen University, Guangzhou, China\\
shaozibo2023@ia.ac.cn, songyg01@pcl.ac.cn, xiaoshan.yang@nlpr.ia.ac.cn
}

\makeatletter
\let\agentpatchmaketitle\maketitle
\let\agentpatchatmaketitle\@maketitle
\let\agentpatchthanks\thanks
\makeatother

\begin{document}

\maketitle

\begin{abstract}
    Agentic multimodal large language models (MLLMs) extend multimodal perception and reasoning with planning, tool use, and interaction in dynamic environments. Yet current models are specialized for particular tools or environments, complicating consolidation into a single generalist. We formulate \textbf{Agentic MLLM Merging} and identify two challenges: asymmetric capability preservation, whereby capabilities with different interaction complexity are retained unevenly, producing weak tasks after merging, and behavior-critical forgetting, whereby losing decisive actions can derail long-horizon execution. We propose \textbf{AgentPatch}, a training-free coarse-to-fine repair framework. It selects a stable merged backbone, restores diluted weak-task-specific signals through Weak-Task Unique Residual Recovery, and applies an Agent-Guided Behavior-Critical Patch that recovers decisive behaviors under explicit capability protection. AgentPatch produces a single static checkpoint without routing or ensembles. Experiments across six agentic and multimodal benchmarks show that AgentPatch improves diverse merged backbones, alleviates weak-task degradation, and better balances weak-task recovery with the preservation of complementary search and agentic visual processing capabilities. Code is available at {\color{blue}\def\UrlFont{\ttfamily}\url{https://github.com/ziboshao/AgentPatch}}.
    \end{abstract}

% Uncomment the following block to link to anonymized code or data.
% \begin{links}
%     \link{Code}{https://anonymous.example.com/code}
% \end{links}

\begin{figure}[!t]
    \centering
    \includegraphics[width=\columnwidth]{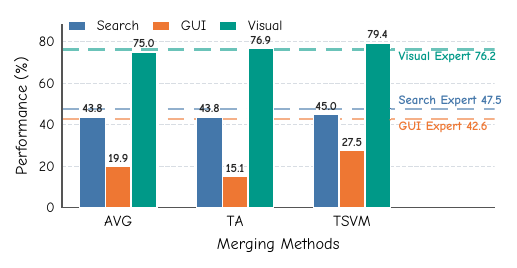}\vspace{-3pt}
    {\small\textbf{(a) Weak-task degradation.}\par}
    \vspace{3pt}
    \includegraphics[width=\columnwidth]{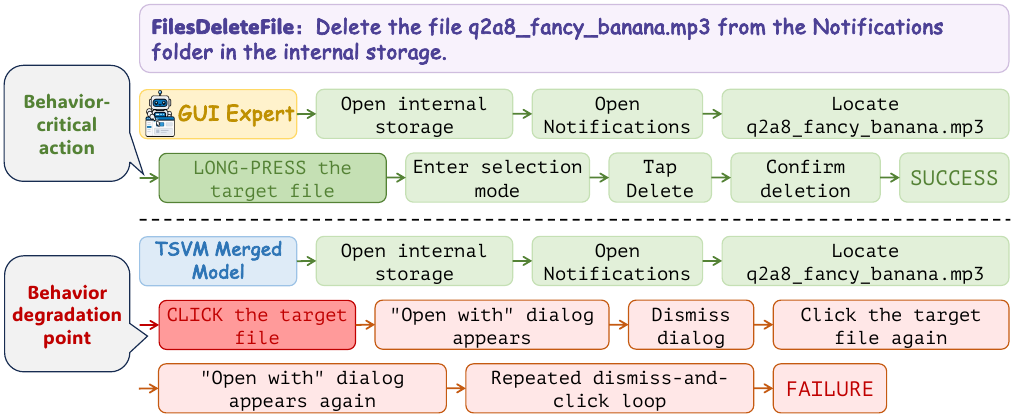}\vspace{2pt}
    {\small\textbf{(b) Behavior-critical degradation.}\par}
    \vspace{-2pt}
    \caption{\textbf{Two challenges of agentic MLLM merging.}
    (a) Results on the development subsets show that merging methods preserve heterogeneous capabilities unevenly, causing severe weak-task degradation.
    (b) Losing a behavior-critical action can change the environment state and trigger trajectory-level failure.}
    \label{fig:agentic-merging-challenges}
\end{figure}

\section{Introduction}

% 直接写agentic mllm
Agentic multimodal large language models (MLLMs) are emerging as autonomous decision-making models that integrate multimodal reasoning, external tool invocation, and interaction with dynamic environments~\cite{yao2025survey}. Building on recent advances in visual perception and cross-modal reasoning~\cite{liu2023visual,li2023blip,zhu2023minigpt,dai2023instructblip,li2024llava,wang2025internvl3,bai2025qwen3}, they can organize multi-step actions, invoke tools when needed, and adapt their decisions according to environmental feedback. These capabilities enable agentic MLLMs to tackle complex tasks involving information seeking, computer operation, and fine-grained visual analysis.

Despite their broad potential, many current agentic MLLMs remain specialized for particular capabilities or interaction environments.

% 可以删掉微调部分的表达
\noindent For example, search agents focus on multimodal information retrieval and evidence synthesis~\cite{wu2025mmsearch,geng2025webwatcher}; GUI agents translate evolving screen observations into grounded actions and multi-step workflows~\cite{ye2025mobile,qin2025ui}; and agentic visual processing models acquire fine-grained evidence through operations such as cropping, zooming, and image manipulation~\cite{zheng2025deepeyes,zhang2025thyme}. Although these specialists achieve strong performance in their target domains, their capabilities often transfer poorly to settings with different action spaces, tool protocols, or trajectory distributions~\cite{feng2026arm}. Deploying separate expert checkpoints also incurs substantial storage, deployment, and system-maintenance overhead.

Model merging~\cite{garipov2018loss,draxler2018essentially,wortsman2022model,choshen2022fusing} provides a cost-effective and modular approach to building generalist agentic MLLMs. By combining expert checkpoints with compatible architectures, typically derived from the same base model, model merging enables the transfer of complementary capabilities into a single model without additional end-to-end training. Existing MLLM merging studies have largely focused on composing static multimodal capabilities under fixed input--output settings, including visual question answering, optical character recognition, chart understanding, visual grounding, multimodal reasoning, modality expansion, and cross-modal alignment~\cite{sung2023empirical,chen2024model,du2025adamms,wei2025unifying,shao2026pivotmerge}. In contrast, agentic MLLMs must operate across heterogeneous observations, tool protocols, action spaces, and long-horizon interaction trajectories, making their capability integration substantially more challenging than static multimodal models. Merging such interactive capabilities within a single MLLM remains underexplored. We therefore formally introduce the \textbf{Agentic MLLM Merging task}, which aims to merge specialized agentic MLLMs developed for different tools and interactive environments into a single generalist model while maintaining balanced performance across diverse agentic tasks.

Agentic MLLM merging makes two challenges particularly salient. First, \textbf{heterogeneous agentic capabilities are not preserved equally}. Their experts capture tasks with different interaction complexity and preservation difficulty, yet global averaging and task-agnostic conflict-resolution criteria do not explicitly account for this asymmetry. Consequently, a merged model may retain some capabilities while substantially diluting another, producing a phenomenon we call \emph{weak-task degradation}. As Figure~\ref{fig:agentic-merging-challenges}(a) illustrates, representative methods, including Weight Averaging~\cite{wortsman2022model}, Task Arithmetic~\cite{ilharco2022editing}, and TSVM~\cite{gargiulo2025task}, retain capabilities unevenly, with the largest expert-to-merge gap appearing on state-dependent, multi-step interaction. Second, \textbf{agentic success is trajectory-sensitive and can hinge on individual behavior-critical decisions}. Unlike a static prediction, each agent action changes subsequent observations, so a single incorrect retrieval, perception, or interaction decision can redirect the remaining trajectory. In Figure~\ref{fig:agentic-merging-challenges}(b), both models locate the target file, but replacing the expert's long-press with a click sends the merged model into an ``Open with'' loop and prevents completion. We call the loss of such decisive patterns \emph{behavior-critical forgetting}. Existing merging methods primarily optimize parameter-level compatibility without explicitly identifying or protecting the localized behaviors that govern trajectory success.

To address these challenges, we propose \textbf{AgentPatch}, a training-free, coarse-to-fine framework for weak-task repair in agentic MLLM merging. AgentPatch first performs \textbf{Stable Backbone Selection} to choose a reliable recipient from candidate training-free merges. To address weak-task degradation, we introduce \textbf{Weak-Task Unique Residual Recovery}, which identifies parameter updates unique to the weak-task expert and selectively restores the signals diluted during merging. Restricting recovery to weak-task-specific coordinates provides broad parameter-level compensation while limiting interference with complementary capabilities. To address behavior-critical forgetting, we further introduce an \textbf{Agent-Guided Behavior-Critical Patch}. It contrasts recipient and expert trajectories to compile repair and protection behaviors, localizes their associated behavior-critical feed-forward neurons, and applies Guardian-constrained expert-directed interpolation. This behavior-level refinement restores decisive execution patterns while preserving complementary agentic and multimodal capabilities.

Our main contributions can be summarized as follows:
\begin{itemize}
    \item We formulate a new \textbf{Agentic MLLM Merging} task, which aims to integrate specialized agentic MLLMs for different tools and interactive environments into a unified generalist model while maintaining balanced performance across diverse agentic tasks.

    \item We propose \textbf{AgentPatch}, a training-free weak-task repair framework for agentic MLLM merging. AgentPatch combines stable backbone selection, weak-task residual recovery, and behavior-critical patching to restore degraded capabilities while reducing interference among different tasks.

    \item We establish systematic evaluation settings across multiple agentic and multimodal benchmarks. Extensive experiments demonstrate that AgentPatch effectively alleviates weak-task degradation and improves the overall balance of merged agentic MLLMs.
\end{itemize}

\section{Related Work}

\subsection{Model Merging}

Model merging integrates multiple expert checkpoints into a single model without unified retraining or additional inference-time cost. Early approaches mainly rely on weight interpolation and task-vector arithmetic. Model Soup~\cite{wortsman2022model} averages fine-tuned checkpoints, while Task Arithmetic~\cite{ilharco2022editing} represents task-specific knowledge as parameter differences and composes capabilities through vector arithmetic. To mitigate interference, TIES-Merging~\cite{yadav2023ties} removes conflicting updates through trimming and sign consensus, whereas DARE~\cite{yu2024language} improves merging via random dropping and rescaling of task-vector elements. Recent methods further improve merging robustness through parameter selection and optimization strategies. Model Stock~\cite{jang2024model} explores interpolation among fine-tuned models, while KnOTS~\cite{stoica2024model} aligns task-vector subspaces for better compatibility. Extending to MLLMs, RobustMerge~\cite{zeng2025robustmerge} and OptMerge~\cite{wei2025unifying} study merging across multimodal tasks and modalities, while PivotMerge~\cite{shao2026pivotmerge} introduces post-alignment merging to integrate cross-modal projectors learned from heterogeneous multimodal pre-training. More recently, ARM~\cite{feng2026arm} extends merging to interactive LLM agents, while complementary LLM-agent research studies execution-time multi-agent workflow adaptation~\cite{xu2026evomas}. However, merging multimodal agents with heterogeneous observations, tools, and long-horizon interaction behaviors remains underexplored.

\subsection{Agentic MLLMs}

Agentic MLLMs combine multimodal perception and reasoning with autonomous decision making, tool use, and sustained interaction with dynamic environments. A recent survey organizes this area along three dimensions: agentic internal intelligence, including reasoning, reflection, and memory; external tool invocation, including search and visual processing; and environment interaction in virtual or physical worlds~\cite{yao2025survey}. Representative systems have developed along specialized capability axes. Agentic search models learn when and how to issue text or image queries and synthesize retrieved evidence~\cite{wu2025mmsearch,geng2025webwatcher}; agentic visual processing models iteratively crop or zoom images to acquire fine-grained evidence~\cite{zheng2025deepeyes,zhang2025thyme}; and GUI agents translate screen observations and interaction histories into grounded actions~\cite{ye2025mobile,qin2025ui}. Although these directions all involve multi-step decision making, they differ substantially in observations, tool protocols, action spaces, supervision, and reward design. Existing work therefore largely optimizes each capability within its own environment. Our work is complementary: rather than training another domain-specific agent, we study how independently developed search, GUI-interaction, and agentic visual processing experts can be consolidated into one agentic MLLM without joint retraining.

\section{Methodology}
\label{sec:methodology}

\begin{figure*}[t]
    \centering
    \includegraphics[width=\textwidth]{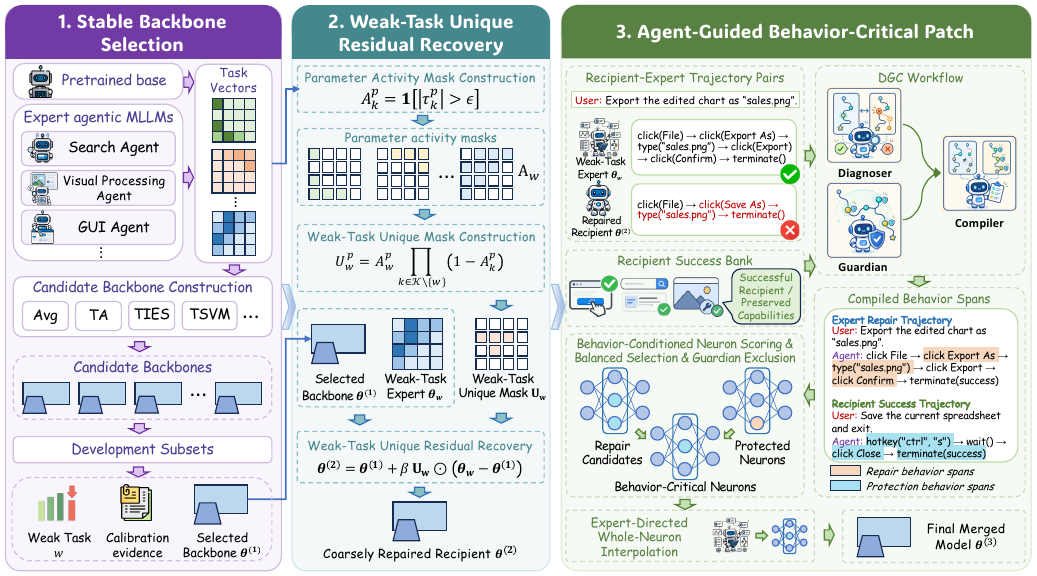}
    \caption{Overview of AgentPatch. Starting from search, GUI-interaction, and agentic visual processing experts derived from a shared base model, AgentPatch first selects a stable merged backbone. It then restores weak-task-exclusive updates diluted during merging through coarse parameter-level residual recovery. Finally, the Diagnoser--Guardian--Compiler (DGC) workflow contrasts recipient and expert trajectories, compiles behavior-critical repair and protection spans, maps them to behavior-critical FFN neurons, and applies a Guardian-constrained expert-directed patch.}
    \label{fig:agentpatch-overview}
\end{figure*}

\subsection{Task Definition}

In this section, we formalize the Agentic MLLM Merging task, including its setting, constraints, and evaluation principle. Let $M_0$ denote a base MLLM with parameters $\boldsymbol{\theta}_0$, and let $\{M_k\}_{k\in\mathcal{K}}$ be a set of expert checkpoints derived from the shared base model. We assume that these experts have compatible architectures and aligned parameterization, while being specialized for heterogeneous agentic capabilities or interactive environments. Each expert induces a task vector
\begin{equation}
    \boldsymbol{\tau}_k
    =
    \boldsymbol{\theta}_k-\boldsymbol{\theta}_0,
\end{equation}
where $\boldsymbol{\theta}_k$ denotes the parameters of expert $M_k$.

Agentic MLLM Merging aims to construct a single generalist checkpoint through a merging procedure $\mathcal{A}$:
\begin{equation}
    \boldsymbol{\theta}^{*}
    =
    \mathcal{A}\!\left(
    \boldsymbol{\theta}_0,
    \{\boldsymbol{\tau}_k\}_{k\in\mathcal{K}}
    \right).
\end{equation}
We consider a training-free setting, where $\mathcal{A}$ introduces no gradient-based optimization or joint multi-environment training. The resulting model is deployed as a single checkpoint without expert routing or ensemble inference.

For each capability $k$, let $\mathcal{E}_k$ denote its associated evaluation environments or benchmarks, and let $S_k(\boldsymbol{\theta})$ be the corresponding aggregate score, with all metrics normalized such that higher values indicate better performance. We define the expert-relative degradation of an arbitrary checkpoint $\boldsymbol{\theta}$ as
\begin{equation}
    D_k(\boldsymbol{\theta})
    =
    \frac{
    S_k(\boldsymbol{\theta}_k)
    -
    S_k(\boldsymbol{\theta})
    }{
    \max\!\left(S_k(\boldsymbol{\theta}_k),\epsilon_s\right)
    }.
\end{equation}
where $\epsilon_s>0$ is a small constant that prevents division by zero. The desired merged model should achieve strong aggregate performance while limiting the largest expert-relative degradation.

Evaluation places the merged model in heterogeneous interactive environments and complementary multimodal benchmarks. The former assess its ability to process evolving observations, select actions, invoke tools, and complete multi-step tasks, while the latter measure whether general multimodal capabilities remain preserved. Accordingly, merging quality is evaluated jointly through capability-specific scores, aggregate performance, and worst-capability degradation.

\subsection{Stable Backbone Selection}

We first construct a stable merged backbone as the recipient for subsequent repair. Different training-free merging operators may preserve expert capabilities unevenly and introduce distinct interference patterns; applying repair to an imbalanced initialization may therefore amplify existing capability gaps. We thus compare candidate backbones produced by multiple merging operators.

Let $\mathcal{G}$ denote the set of candidate merging operators. Each operator $\mathcal{G}_g$, $g\in\mathcal{G}$, produces a candidate from the shared base and expert task vectors:
\begin{equation}
\boldsymbol{\theta}^{(g)}
=
\mathcal{G}_g
\left(
\boldsymbol{\theta}_0,
\{\boldsymbol{\tau}_k\}_{k\in\mathcal{K}}
\right),
\end{equation}
Candidate selection is performed on fixed development subsets covering all target capabilities, while the test sets remain unseen. Rather than optimizing an individual capability, we select the candidate with the highest unweighted mean of normalized capability scores:
\begin{equation}
g^{*}
=
\arg\max_{g\in\mathcal{G}}
\frac{1}{|\mathcal{K}|}
\sum_{k\in\mathcal{K}}
S_k\!\left(\boldsymbol{\theta}^{(g)}\right),
\qquad
\boldsymbol{\theta}^{(1)}
=
\boldsymbol{\theta}^{(g^{*})},
\end{equation}
We then identify its weakest-preserved capability on the same subsets as
\begin{equation}
w
=
\arg\max_{k\in\mathcal{K}}
D_k\!\left(\boldsymbol{\theta}^{(1)}\right).
\end{equation}

This stage provides both the recipient $\boldsymbol{\theta}^{(1)}$ and the target capability $w$ for the subsequent coarse-to-fine repair.

\subsection{Weak-Task Unique Residual Recovery}

Although the selected backbone provides a stable starting point, it may still preserve some expert capabilities substantially less effectively than others. Directly injecting the complete task vector of the weakest-preserved expert may recover task-specific signals, but can also perturb parameter coordinates modified by other experts, thereby introducing negative transfer. We therefore perform a coarse parameter-level repair restricted to updates unique to the weak-task expert.

Let $w\in\mathcal{K}$ denote the weakest-preserved capability identified on the calibration set. For each scalar parameter coordinate $p$, we define the activity indicator of expert $k$ as
\begin{equation}
    A_k^{p}
    =
    \mathbf{1}\!\left[
    \left|\tau_k^{p}\right|>\epsilon
    \right],
    \qquad k\in\mathcal{K},
\end{equation}
where $\epsilon$ filters negligible parameter variations. The weak-task unique mask is then defined as
\begin{equation}
    U_w^{p}
    =
    A_w^{p}
    \prod_{k\in\mathcal{K}\setminus\{w\}}
    \left(1-A_k^{p}\right).
\end{equation}
This mask retains coordinates modified by the weak-task expert while excluding those that are also active in any other expert.

We recover the missing weak-task residual by interpolating the selected backbone toward the corresponding expert only on these coordinates:
\begin{equation}
    \boldsymbol{\theta}^{(2)}
    =
    \boldsymbol{\theta}^{(1)}
    +
    \beta\,\mathbf{U}_w\odot
    \left(
    \boldsymbol{\theta}_w
    -
    \boldsymbol{\theta}^{(1)}
    \right),
    \label{eq:coarse-repair}
\end{equation}
where $\beta$ controls the recovery strength and $\odot$ denotes element-wise multiplication. Unlike directly adding the complete expert task vector, this operation restores only the expert-directed residual on weak-task-exclusive coordinates, providing broad capability compensation while limiting interference with the remaining experts. As a coarse parameter-level repair, however, it does not distinguish which local actions or behavior patterns are decisive for successful interaction, motivating the subsequent behavior-critical patch.

\subsection{Agent-Guided Behavior-Critical Patch}

The coarse repair cannot distinguish behavior patterns that are decisive for successful interaction. The final stage therefore identifies trajectory gaps, traces the corresponding feed-forward neurons, excludes neurons associated with protected capabilities, and softly interpolates the remaining neurons toward the weak-task expert.

\subsubsection{Agent-Guided Behavior Specification}

For each calibration task, we collect paired trajectories generated on the same task by the repaired recipient $\boldsymbol{\theta}^{(2)}$ and the weak-task expert $\boldsymbol{\theta}_w$; all calibration tasks are disjoint from the test tasks. The \emph{Diagnoser} compares each pair and identifies behavior gaps for which the recipient fails while the expert succeeds. The corresponding expert behaviors form the \emph{repair evidence}.

To reduce collateral degradation, the \emph{Guardian} gathers \emph{protection evidence} from behaviors that the recipient already performs correctly, including successful interaction patterns, valid action formats, and capabilities inherited from other experts. The \emph{Compiler} then converts the repair and protection evidence into token-level span selectors while preserving their environment and trajectory-step provenance. Together, these three analysis agents form the Diagnoser--Guardian--Compiler (DGC) workflow and produce two collections of behavior spans: desired expert spans to be restored and recipient spans whose associated functionality should be protected. DGC is used only during offline checkpoint construction to produce auditable token-span selectors and introduces no analysis-agent dependency at inference time.

\subsubsection{Behavior-Conditioned Neuron Scoring}

We apply teacher forcing to the fixed behavior trajectories and record feed-forward activations only at the compiled token positions. Let $\mathbf{z}^{\ell}_t$ denote the intermediate FFN activation of width $d_{\mathrm{ff}}$ at token $t$ in layer $\ell$. The output contributed by neuron $j$ is $z^{\ell}_{t,j}\mathbf{W}^{\ell}_{\mathrm{out}}[:,j]$.

Let $\mathcal{D}^{\mathrm{rep}}_r$ denote the expert trajectories associated with repair group $r$, and let
$\mathcal{T}^{\mathrm{rep}}_{e,r}$ be the selected behavior-token positions in trajectory $e$. We compute the trajectory-balanced activation of neuron $j$ as
\begin{equation}
    a^{\mathrm{rep}}_{\ell rj}
    =
    \frac{1}{|\mathcal{D}^{\mathrm{rep}}_r|}
    \sum_{e\in\mathcal{D}^{\mathrm{rep}}_r}
    \frac{1}{|\mathcal{T}^{\mathrm{rep}}_{e,r}|}
    \sum_{t\in\mathcal{T}^{\mathrm{rep}}_{e,r}}
    \left|z^{\ell,w}_{t,j}\right|,
    \label{eq:abs-activation}
\end{equation}
where $z^{\ell,w}_{t,j}$ is obtained from the weak-task expert. Averaging first within each trajectory prevents longer action sequences from dominating the estimate.

Activation magnitude alone does not reflect how strongly a neuron affects the layer output. We therefore incorporate the norm of its outgoing projection and define
\begin{equation}
    c^{\mathrm{rep}}_{\ell rj}
    =
    a^{\mathrm{rep}}_{\ell rj}
    \left\|
    \mathbf{W}^{\ell,w}_{\mathrm{out}}[:,j]
    \right\|_2.
    \label{eq:output-contribution}
\end{equation}
This score favors neurons that are both active on the desired behavior and capable of substantially influencing downstream representations. For each repair group $r$, we retain a group-level fraction $q_r$ of the highest-scoring neurons in every layer:
\begin{equation}
    \mathcal{R}^{\ell}_r
    =
    \operatorname{TopK}\!\left(
    \{c^{\mathrm{rep}}_{\ell rj}\}_{j=1}^{d_{\mathrm{ff}}},
    \left\lceil q_r d_{\mathrm{ff}}\right\rceil
    \right).
    \label{eq:repair-topk}
\end{equation}

\subsubsection{Balanced Selection and Guardian Exclusion}

Calibration environments may contain different numbers of trajectories or behavior groups, so a global selection budget can favor evidence-rich environments. We therefore select neurons independently within each group using $q_r$ before taking their union, maintaining balanced contributions across heterogeneous environments.

Protection neurons are identified analogously from protected recipient spans. Specifically, activations and output contributions are computed using the recipient
$\boldsymbol{\theta}^{(2)}$. Let $c^{\mathrm{prot}}_{\ell pj}$ denote the resulting contribution score for protection group $p$. With a protection fraction $q_{\mathrm{guard}}$, we define
\begin{equation}
    \mathcal{P}^{\ell}_p
    =
    \operatorname{TopK}\!\left(
    \{c^{\mathrm{prot}}_{\ell pj}\}_{j=1}^{d_{\mathrm{ff}}},
    \left\lceil q_{\mathrm{guard}} d_{\mathrm{ff}}\right\rceil
    \right).
    \label{eq:protection-topk}
\end{equation}
The final repair set is
\begin{equation}
    \mathcal{M}^{\ell}
    =
    \left(
    \bigcup_r \mathcal{R}^{\ell}_r
    \right)
    \setminus
    \left(
    \bigcup_p \mathcal{P}^{\ell}_p
    \right).
    \label{eq:guardian}
\end{equation}
We refer to the surviving neurons in $\mathcal{M}^{\ell}$ as \emph{behavior-critical neurons}. A neuron associated with any protected behavior is excluded from editing, even when it also receives a high repair score. This conservative Guardian mechanism prioritizes capability preservation over aggressive weak-task recovery.

\subsubsection{Expert-Directed Whole-Neuron Interpolation}

Let $\boldsymbol{\phi}^{\ell}_{j}$ denote the complete parameter block associated with neuron $j$, including its incoming and outgoing connections. For every selected neuron $j\in\mathcal{M}^{\ell}$, we update
\begin{equation}
    \boldsymbol{\phi}^{\ell,(3)}_{j}
    =
    (1-\alpha)
    \boldsymbol{\phi}^{\ell,(2)}_{j}
    +
    \alpha
    \boldsymbol{\phi}^{\ell,w}_{j}.
    \label{eq:neuron-write}
\end{equation}
All unselected parameters remain unchanged. Unlike hard neuron replacement, this soft expert-directed interpolation preserves most of the recipient representation while selectively restoring behavior-critical expert signals.

\section{Experiments}

\begin{table*}[t]
\centering
\normalsize
\setlength{\tabcolsep}{5.0pt}
\begin{tabular}{lccccccc}
\toprule
Method & MMSearch & FactualVQA & AndroidWorld & OSWorld & V*Bench & HR-Bench 8K & Avg. \\
\midrule
\rowcolor{tablegroupgray}
\multicolumn{8}{l}{\textit{Expert agentic MLLMs}} \\
MMSearch-R1-7B & 47.0 & 52.8 & 30.4 & 13.3 & 72.2 & 68.3 & 47.4 \\
GUI-Owl-7B & 17.2 & 24.9 & 71.2 & 28.5 & 34.4 & 62.6 & 39.8 \\
DeepEyes-7B & 29.8 & 33.1 & 16.3 & 12.4 & 84.1 & 70.0 & 41.0 \\
\midrule
\rowcolor{tablegroupgray}
\multicolumn{8}{l}{\textit{Merging Methods}} \\
Weight Averaging & 39.1 & 45.6 & 46.7 & \underline{19.0} & 75.5 & 69.6 & 49.3 \\
Task Arithmetic & 41.1 & 43.1 & 43.5 & 15.4 & 76.2 & 69.1 & 48.0 \\
TIES-Merging & 0.0 & 7.4 & 44.6 & \textbf{19.1} & 2.0 & 61.2 & 22.4 \\
TSVM & \underline{44.4} & 49.6 & \underline{59.2} & 17.7 & \underline{84.8} & 71.1 & \underline{54.5} \\
Iso-CTS & 40.4 & 50.9 & 43.5 & 11.2 & 83.4 & \textbf{75.8} & 50.9 \\
OptMerge & 39.7 & 47.7 & 38.0 & 11.5 & 74.8 & 70.0 & 47.0 \\
ACE-Merging & 41.7 & \underline{52.4} & 44.6 & 11.9 & \textbf{85.4} & \underline{73.8} & 51.6 \\
DC-Merge & 39.7 & 36.2 & 50.0 & 15.6 & 78.2 & 69.7 & 48.2 \\
\rowcolor{tablemethodblue}
\textbf{AgentPatch} & \textbf{46.4} & \textbf{52.7} & \textbf{63.6} & \underline{19.0} & \underline{84.8} & 73.1 & \textbf{56.6} \\
\bottomrule
\end{tabular}
\caption{Test-set performance across agentic and multimodal benchmarks. All values are percentages, and higher is better. Bold and underlined numbers indicate the best and second-best results, respectively, among merging methods with an available score. The overall score is reported only when all six task results are available.}
\label{tab:main-results}
\end{table*}

\subsection{Experimental Setup}

\paragraph{Models.}
We conduct all experiments with Qwen2.5-VL-7B~\cite{bai2025qwen25vltechnicalreport} as the shared base architecture. The three source experts are MMSearch-R1-7B for agentic multimodal search~\cite{wu2025mmsearch}, GUI-Owl-7B for GUI interaction~\cite{ye2025mobile}, and DeepEyes-7B for agentic visual processing~\cite{zheng2025deepeyes}. AgentPatch produces one static checkpoint and uses neither expert routing nor ensemble inference at test time.

\paragraph{Benchmarks.}
We evaluate six benchmarks covering three task types. MMSearch~\cite{jiang2024mmsearch} and FactualVQA~\cite{wu2025mmsearch} assess agentic multimodal search; AndroidWorld~\cite{rawles2024androidworld} and OSWorld~\cite{xie2024osworld} evaluate GUI interaction in mobile and desktop environments; V*Bench~\cite{wu2023vstar} and HR-Bench 8K~\cite{wang2024dc2} assess agentic visual processing through fine-grained visual search and high-resolution image understanding.

\paragraph{Baselines.}
We compare AgentPatch with representative training-free merging methods, including Weight Averaging~\cite{wortsman2022model}, Task Arithmetic~\cite{ilharco2022editing}, TIES-Merging~\cite{yadav2023ties}, TSVM~\cite{gargiulo2025task}, Iso-CTS~\cite{marczak2025notaskleftbehind}, OptMerge~\cite{wei2025unifying}, ACE-Merging~\cite{xu2026ace}, and DC-Merge~\cite{zhang2026dcmerge}.

\paragraph{Implementation details.}
We construct candidate backbones using Weight Averaging, Task Arithmetic, TIES-Merging, and TSVM, and select TSVM on the fixed calibration partitions. The same calibration results identify GUI interaction as the weakest-preserved capability, which is targeted by the subsequent repair stages. The activity threshold, unique-residual coefficient, and per-group Guardian protection ratio are set to $\epsilon=10^{-5}$, $0.15$, and $0.01$, respectively. We interpolate the selected behavior-critical neurons toward GUI-Owl-7B with coefficient $0.02$. No gradients or test examples are used during checkpoint construction.

\subsection{Main Results}

Table~\ref{tab:main-results} first confirms the complementary specialization of the three source experts. MMSearch-R1-7B performs best among the experts on both search benchmarks, GUI-Owl-7B leads on both GUI benchmarks, and DeepEyes-7B leads on both agentic visual processing benchmarks. Each expert's advantage is concentrated in its target task type, motivating the consolidation of their complementary capabilities into a single agentic MLLM.

Conventional merging improves coverage but does not reliably resolve this imbalance. Weight Averaging and Task Arithmetic obtain overall scores of 49.3 and 48.0, respectively. Among the stronger structure-aware baselines, TSVM provides the most balanced initialization, reaching 54.5 on average, while other approaches attain competitive results on individual visual benchmarks but remain weaker on GUI interaction.

AgentPatch achieves the best complete overall result, improving the TSVM backbone from 54.5 to 56.6. It improves MMSearch, FactualVQA, AndroidWorld, OSWorld, and HR-Bench 8K while preserving V*Bench performance. In particular, the gains on both AndroidWorld and OSWorld show that weak-task repair transfers across mobile and desktop interaction environments rather than overfitting to a single GUI benchmark. At the same time, the model remains close to the search expert on MMSearch and FactualVQA and to the visual expert on V*Bench, demonstrating a better balance between weak-task recovery and preservation of complementary capabilities.

\subsection{Ablation Studies}

\begin{table}[t]
\centering
\small
\begin{tabular*}{\columnwidth}{@{\extracolsep{\fill}}lcccc@{}}
\toprule
Variant & Search & GUI & Visual & Avg. \\
\midrule
TSVM Backbone & 47.0 & 38.5 & 78.0 & 54.5 \\
$+$ Unique Residual & \textbf{51.0} & 39.8 & 77.7 & 56.2 \\
$+$ Behavior Patch & 49.6 & \textbf{41.3} & \textbf{79.0} & \textbf{56.6} \\
\bottomrule
\end{tabular*}
\caption{Component ablation by task type. Search averages MMSearch and FactualVQA; GUI averages AndroidWorld and OSWorld; Visual averages V*Bench and HR-Bench 8K. Avg. is the mean over all six benchmarks.}
\label{tab:component-ablation}
\end{table}

\begin{figure}[t]
\centering
\includegraphics[width=\columnwidth]{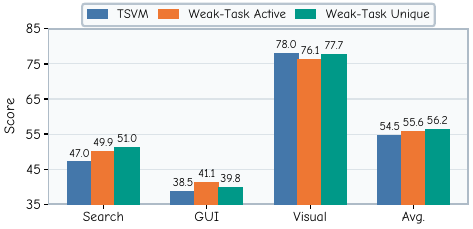}
\caption{Effect of weak-task residual support relative to the TSVM backbone. Weak-Task Active uses all coordinates active in the weak-task expert, whereas Weak-Task Unique retains only coordinates inactive in the other experts. Each task type averages the same two test benchmarks as in Table~\ref{tab:component-ablation}; Avg. is the mean over all six benchmarks.}
\label{fig:agentpatch-ablation}
\end{figure}

\begin{table}[t]
\centering
\small
\begin{tabular*}{\columnwidth}{@{\extracolsep{\fill}}lcccc@{}}
\toprule
Variant & Search & GUI & Visual & Avg. \\
\midrule
Random Neurons & \textbf{50.1} & 35.7 & 77.4 & \underline{54.4} \\
w/o Guardian & 47.9 & \underline{36.3} & \underline{78.4} & 54.2 \\
Full AgentPatch & \underline{49.6} & \textbf{41.3} & \textbf{79.0} & \textbf{56.6} \\
\bottomrule
\end{tabular*}
\caption{Core design ablation for the behavior-critical patch. Random Neurons replaces behavior-conditioned localization with layer-matched random selection, while w/o Guardian disables protected-neuron exclusion.}
\label{tab:behavior-patch-ablation}
\end{table}

\paragraph{Component-wise contributions.}
Table~\ref{tab:component-ablation} evaluates the two repair stages on the selected TSVM backbone. Weak-Task Unique Residual Recovery raises the overall average from 54.5 to 56.2 while improving Search from 47.0 to 51.0 and GUI from 38.5 to 39.8. The behavior-critical patch further raises GUI from 39.8 to 41.3 and Visual from 77.7 to 79.0, yielding the best average of 56.6. These complementary gains support the coarse-to-fine design: parameter-level recovery broadly restores weak-task signals, while behavior-level refinement strengthens decisive interaction patterns and preserves the integrated capability balance.

\paragraph{Effect of unique residual support.}
Figure~\ref{fig:agentpatch-ablation} compares all active coordinates of the weak-task expert with its unique subset while keeping the backbone and residual strength fixed. Although the active region raises GUI from 39.8 to 41.1, it reduces Search and Visual by 1.1 and 1.6 points and lowers the overall average from 56.2 to 55.6. The weak-task-unique mask therefore does more than sparsify the update: excluding coordinates also active in other experts yields a better balance between weak-task recovery and preservation of complementary capabilities.

\paragraph{Effect of behavior localization and capability protection.}
Table~\ref{tab:behavior-patch-ablation} disentangles the two core designs in the behavior-critical patch. Replacing behavior-conditioned localization with layer-matched random neurons reduces GUI by 5.6 points and the overall average by 2.2 points despite a higher Search score, showing that the gain does not arise from arbitrary sparse writing. Disabling Guardian exclusion similarly lowers GUI by 5.0 points, Visual by 0.6 points, and the overall average by 2.4 points. The full design achieves the strongest balance, demonstrating the complementary roles of behavior-critical localization and Guardian-based protection in weak-task repair and capability preservation.

\subsection{Analysis}

\begin{figure}[t]
\centering
\includegraphics[width=\columnwidth]{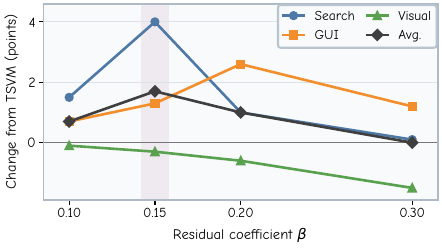}
\caption{Sensitivity to the unique-residual coefficient $\beta$, reported as performance change from the TSVM backbone.}
\label{fig:agentpatch-analysis}
\end{figure}

\paragraph{Sensitivity to residual strength.}
Figure~\ref{fig:agentpatch-analysis} varies only the interpolation coefficient $\beta$ in Equation~\ref{eq:coarse-repair}. The intermediate value $\beta=0.15$ achieves the best overall balance, with the highest Search average and an overall score of 56.2. Increasing $\beta$ to 0.20 further improves GUI, but stronger residual injection shifts the balance away from Search and Visual. At $\beta=0.30$, the overall score returns to that of the original backbone. The trend confirms that weak-task recovery is not monotonic: moderate interpolation restores task-specific signals, whereas overly strong injection increasingly interferes with complementary capabilities.

\begin{table}[t]
\centering
\small
\begin{tabular*}{\columnwidth}{@{\extracolsep{\fill}}lcc@{}}
\toprule
Recipient & GUI & Avg. \\
\midrule
TSVM & $38.5 \rightarrow \mathbf{41.3}$ & $54.5 \rightarrow \mathbf{56.6}$ \\
Iso-CTS & $27.3 \rightarrow \mathbf{30.0}$ & $50.9 \rightarrow \mathbf{52.0}$ \\
\bottomrule
\end{tabular*}
\caption{Generalization across merged recipients. Each cell reports the task-type or six-benchmark average before and after applying AgentPatch. GUI averages AndroidWorld and OSWorld.}
\label{tab:recipient-generalization}
\end{table}

\paragraph{Generalization across merged recipients.}
Table~\ref{tab:recipient-generalization} applies the same AgentPatch pipeline to TSVM and Iso-CTS, with behavior-critical neurons localized separately for each recipient. For TSVM, AgentPatch improves the GUI average by 2.8 points and the overall six-benchmark average by 2.1 points; for Iso-CTS, the corresponding gains are 2.7 and 1.1 points. The consistent improvements across recipients show that the repair is not tied to a single backbone construction rule.

\begin{figure}[t]
\centering
\includegraphics[width=\columnwidth]{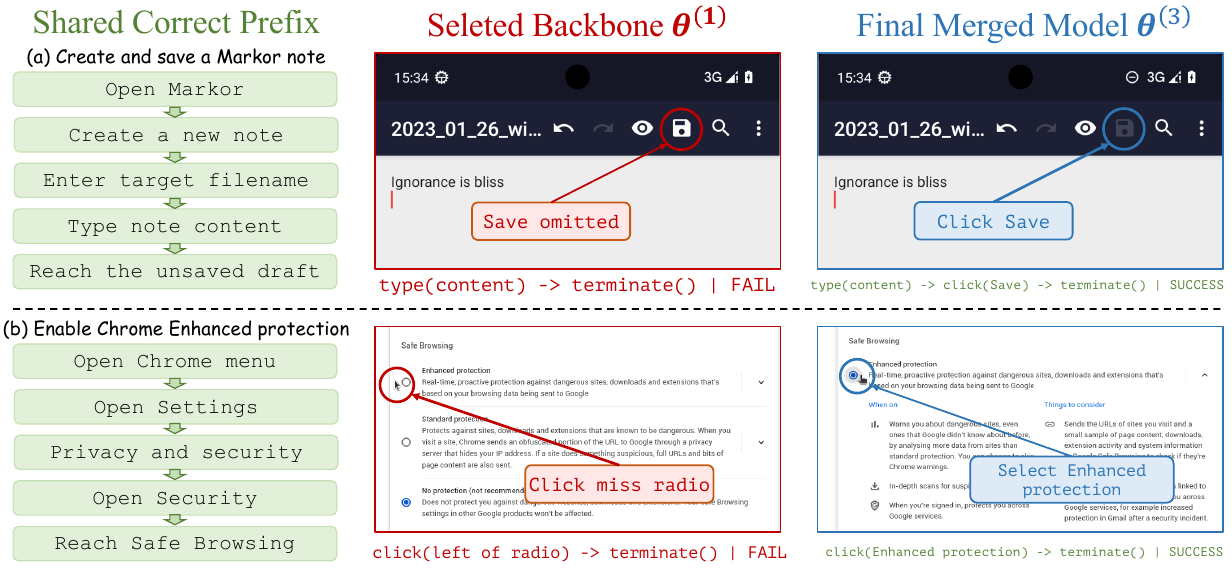}
\caption{Qualitative examples of behavior-critical recovery. Both models execute the shared correct prefix, but the selected backbone omits or misgrounds a decisive action. AgentPatch recovers the action and completes the task.}
\label{fig:case-study}
\end{figure}

\paragraph{Case study of behavior-critical recovery.}
Figure~\ref{fig:case-study} illustrates how a single localized action can determine the outcome of an otherwise correct trajectory. In the Markor example, both models create the target note and enter its content, but the selected backbone terminates without saving it; AgentPatch restores the missing \texttt{click(Save)} action and completes the task. In the Chrome example, both models reach the Safe Browsing page, yet the selected backbone clicks beside the Enhanced Protection radio button. AgentPatch grounds the action on the intended control and succeeds. These cases show that AgentPatch preserves shared correct prefixes while recovering the behavior-critical actions that turn near-complete trajectories into successful executions.

\section{Conclusion}

This paper introduced Agentic MLLM Merging, a new setting for consolidating specialists that operate with heterogeneous tools, observations, and interaction environments. We identified two central challenges in this setting: weak-task degradation caused by asymmetric capability preservation and the loss of behavior-critical patterns that determine long-horizon task success. To address them, we proposed AgentPatch, a training-free, coarse-to-fine framework that selects a stable merged backbone, restores weak-task-exclusive residuals, and applies an agent-guided behavior-critical patch with explicit capability protection. Experiments across six agentic and multimodal benchmarks show that AgentPatch improves the overall balance of the merged model, strengthens weak interactive capabilities across mobile and desktop environments, and better preserves complementary search and agentic visual processing skills. These findings show that effective agentic model consolidation requires more than resolving global parameter conflicts: task-specific residuals and localized behaviors governing trajectory success must also be recovered. This yields a deployable generalist checkpoint without joint retraining or inference-time routing.

\bibliography{aaai2027}

\clearpage

% Supplementary-material-only definitions and layout settings.
\definecolor{dgcrepair}{RGB}{252,235,219}
\definecolor{dgcprotect}{RGB}{221,239,248}
\definecolor{dgcneutral}{RGB}{239,244,240}
\colorlet{dgcrepairdark}{dgcrepair!58!black}
\colorlet{dgcprotectdark}{dgcprotect!58!black}
\colorlet{dgcneutraldark}{dgcneutral!58!black}
\newenvironment{promptbox}[3]{%
  \def\FrameCommand##1{\fcolorbox{#2}{#3}{##1}}%
  \MakeFramed{\advance\hsize-\width\FrameRestore}%
  \noindent\colorbox{#2}{\parbox{\dimexpr\linewidth-2\fboxsep\relax}{%
    \color{white}\bfseries #1}}\par
  \smallskip
  \scriptsize\ttfamily\raggedright
  \setlength{\parindent}{0pt}%
  \setlength{\parskip}{3pt}%
}{\endMakeFramed}
\setlength{\textfloatsep}{6pt plus 2pt minus 2pt}
\setlength{\floatsep}{6pt plus 2pt minus 2pt}
\setlength{\intextsep}{6pt plus 2pt minus 2pt}
\setlength{\abovecaptionskip}{3pt}
\setlength{\belowcaptionskip}{2pt}

\setcounter{section}{0}
\setcounter{subsection}{0}
\setcounter{figure}{0}
\setcounter{table}{0}
\setcounter{algorithm}{0}
\setcounter{equation}{0}
\setcounter{page}{1}
\setcounter{secnumdepth}{2}
\renewcommand{\thesection}{\Alph{section}}
\renewcommand{\thesubsection}{\thesection.\arabic{subsection}}

\makeatletter
\let\maketitle\agentpatchmaketitle
\let\@maketitle\agentpatchatmaketitle
\let\thanks\agentpatchthanks
% The AAAI title macro allocates these helpers internally.  Release only
% those helpers so the unchanged supplementary title can be typeset again.
\let\c@aaai@eqfn\@undefined
\let\c@aaai@corrfn\@undefined
\let\titlearea\@undefined
\let\actualheight\@undefined
\makeatother
\title{Supplementary Material for\\
\textit{AgentPatch: Coarse-to-Fine Weak-Task Repair for Merging Agentic Multimodal Large Language Models}}
\author{
Zibo Shao\textsuperscript{\rm 1,\rm 2,\rm 3},
Baochen Xiong\textsuperscript{\rm 1,\rm 2,\rm 3},
Chengdong Xu\textsuperscript{\rm 4},
Linhui Xiao\textsuperscript{\rm 2},\\
Kaichen Li\textsuperscript{\rm 1,\rm 2,\rm 3},
Haoran Gong\textsuperscript{\rm 2},
Yan Li\textsuperscript{\rm 2},
Yaguang Song\textsuperscript{\rm 2}\corresponding,
Xiaoshan Yang\textsuperscript{\rm 1,\rm 2,\rm 3}\corresponding
}
\affiliations{
\textsuperscript{\rm 1}State Key Laboratory of Multimodal Artificial Intelligence Systems, Institute of Automation,\\
Chinese Academy of Sciences, Beijing, China\\
\textsuperscript{\rm 2}Pengcheng Laboratory, Shenzhen, China\\
\textsuperscript{\rm 3}School of Artificial Intelligence, University of Chinese Academy of Sciences, Beijing, China\\
\textsuperscript{\rm 4}Sun Yat-Sen University, Guangzhou, China\\
shaozibo2023@ia.ac.cn, songyg01@pcl.ac.cn, xiaoshan.yang@nlpr.ia.ac.cn
}
\maketitle

\section{Additional Method Details}
\label{sec:supp-method}

\subsection{Overall Algorithm}
\label{sec:supp-overall-algorithm}

\begin{algorithm}[H]
\small
\caption{AgentPatch: Training-Free Coarse-to-Fine Weak-Task Repair}
\label{alg:agentpatch}
\begin{algorithmic}[1]
\REQUIRE Base $\boldsymbol{\theta}_0$; aligned experts
$\{\boldsymbol{\theta}_k\}_{k\in\mathcal{K}}$; merging operators
$\mathcal{G}$; calibration partitions $\{\mathcal{C}_k\}_{k\in\mathcal{K}}$
\ENSURE Final merged checkpoint $\boldsymbol{\theta}^{(3)}$
\STATE \textbf{Stage 1: Stable backbone selection}
\STATE $\boldsymbol{\tau}_k\gets
\boldsymbol{\theta}_k-\boldsymbol{\theta}_0,\ \forall k\in\mathcal{K}$
\FOR{each merging operator $g\in\mathcal{G}$}
    \STATE $\boldsymbol{\theta}^{(g)}\gets
    \mathcal{G}_g(\boldsymbol{\theta}_0,
    \{\boldsymbol{\tau}_k\}_{k\in\mathcal{K}})$
    \STATE $\bar S_g\gets |\mathcal{K}|^{-1}
    \sum_{k\in\mathcal{K}}S_k(\boldsymbol{\theta}^{(g)})$
\ENDFOR
\STATE $g^*\gets\arg\max_g\bar S_g$,
$\boldsymbol{\theta}^{(1)}\gets\boldsymbol{\theta}^{(g^*)}$
\STATE $w\gets\arg\max_kD_k(\boldsymbol{\theta}^{(1)})$
\STATE \textbf{Stage 2: Weak-task unique residual recovery}
\STATE $U_w^p\gets \mathbf{1}[|\tau_w^p|>\epsilon]
\prod_{k\ne w}\mathbf{1}[|\tau_k^p|\leq\epsilon]$
\STATE $\boldsymbol{\theta}^{(2)}\gets\boldsymbol{\theta}^{(1)}+
\beta\mathbf{U}_w\odot
(\boldsymbol{\theta}_w-\boldsymbol{\theta}^{(1)})$
\STATE \textbf{Stage 3: Agent-guided behavior-critical patch}
\STATE Collect paired weak-task trajectories $\mathcal{H}^{\mathrm{rep}}$
and successful recipient trajectories $\mathcal{H}^{\mathrm{prot}}$.
\STATE $(\mathcal{S}^{\mathrm{rep}},\mathcal{S}^{\mathrm{prot}})
\gets\operatorname{DGC}(\mathcal{H}^{\mathrm{rep}},
\mathcal{H}^{\mathrm{prot}})$
\STATE $c^{\mathrm{rep}}_{\ell rj}\gets a^{\mathrm{rep}}_{\ell rj}
\|\mathbf{W}^{\ell,w}_{\mathrm{out}}[:,j]\|_2$
\STATE $c^{\mathrm{prot}}_{\ell pj}\gets a^{\mathrm{prot}}_{\ell pj}
\|\mathbf{W}^{\ell,(2)}_{\mathrm{out}}[:,j]\|_2$
\STATE $\mathcal{R}^{\ell}\gets\bigcup_r
\operatorname{TopK}(c^{\mathrm{rep}}_{\ell r:},
\lceil q_r d_{\mathrm{ff}}\rceil)$
\STATE $\mathcal{P}^{\ell}\gets\bigcup_p
\operatorname{TopK}(c^{\mathrm{prot}}_{\ell p:},
\lceil q_{\mathrm{guard}}d_{\mathrm{ff}}\rceil)$
\STATE $\mathcal{M}^{\ell}\gets
\mathcal{R}^{\ell}\setminus\mathcal{P}^{\ell}$
\STATE $\boldsymbol{\theta}^{(3)}\gets\boldsymbol{\theta}^{(2)}$
\FOR{each $j\in\mathcal{M}^{\ell}$ and layer $\ell$}
    \STATE $\boldsymbol{\phi}^{\ell,(3)}_j\gets
    (1-\alpha)\boldsymbol{\phi}^{\ell,(2)}_j+
    \alpha\boldsymbol{\phi}^{\ell,w}_j$
\ENDFOR
\RETURN $\boldsymbol{\theta}^{(3)}$
\end{algorithmic}
\end{algorithm}

Algorithm~\ref{alg:agentpatch} gives an executable summary of the complete
AgentPatch pipeline. In particular, backbone selection produces the recipient
$\boldsymbol{\theta}^{(1)}$ and the weakest-preserved capability $w$, which
jointly determine the subsequent repair target. DGC is invoked only during
offline checkpoint construction, so the procedure outputs a single static
checkpoint with no analysis-agent, expert-routing, or ensemble dependency at
inference time.

\subsection{Development and Test Partitions}
\label{sec:supp-data-partitions}

Table~\ref{tab:supp-data-partitions} specifies the fixed partitions used
throughout the paper. For each benchmark, the development partition supports
candidate selection, weak-capability identification, or offline repair
construction, whereas the reported test partition is its disjoint
complement.

\begin{table}[t]
\centering
\small
\setlength{\tabcolsep}{1.9pt}
\renewcommand{\arraystretch}{1.10}
\begin{tabular*}{\columnwidth}{@{\extracolsep{\fill}}lrrrrr@{}}
\toprule
\textbf{Benchmark} & \textbf{Pool} & \textbf{Dev.} & \textbf{Test} &
\textbf{Dev./Pool} & \textbf{Dev./Test} \\
\midrule
MMSearch              & 171 & 20 & 151 & 11.7\% & 13.2\% \\
FactualVQA            & 1,800 & 120 & 1,680 & 6.7\% & 7.1\% \\
AndroidWorld          & 116 & 24 & 92 & 20.7\% & 26.1\% \\
OSWorld               & 369 & 74 & 295 & 20.1\% & 25.1\% \\
V*Bench               & 191 & 40 & 151 & 20.9\% & 26.5\% \\
HR-Bench 8K           & 800 & 80 & 720 & 10.0\% & 11.1\% \\
\bottomrule
\end{tabular*}
\caption{Development and test partitions. ``Pool'' denotes
the complete evaluated benchmark pool. We report both the development share
of the pool and its size relative to the test complement.}
\label{tab:supp-data-partitions}
\end{table}

\subsection{Stable Backbone Selection Details}
\label{sec:supp-backbone-selection}

We instantiate the candidate set $\mathcal{G}$ with Weight Averaging, Task
Arithmetic, TIES-Merging, and TSVM. All four candidates use the same base and
aligned expert checkpoints and are evaluated with identical inference protocols
on fixed calibration partitions. No test example participates in candidate
selection or weak-capability identification.

For capability $k$, let $\mathcal{B}_k$ be its associated calibration
benchmarks and let $s_b(\boldsymbol{\theta})\in[0,1]$ denote the score of model
$\boldsymbol{\theta}$ on benchmark $b$ after metric normalization. We aggregate
the benchmarks belonging to the same capability as
\begin{equation}
S_k(\boldsymbol{\theta})
=
\frac{1}{|\mathcal{B}_k|}
\sum_{b\in\mathcal{B}_k}s_b(\boldsymbol{\theta}).
\label{eq:supp-capability-score}
\end{equation}
Following the main paper, with $\epsilon_s>0$ denoting a small constant that
prevents division by zero, we measure the preservation of capability $k$ by
its expert-relative degradation
\begin{equation}
D_k(\boldsymbol{\theta})
=
\frac{S_k(\boldsymbol{\theta}_k)-S_k(\boldsymbol{\theta})}
{\max\!\left(S_k(\boldsymbol{\theta}_k),\epsilon_s\right)},
\label{eq:supp-expert-relative-degradation}
\end{equation}
Because each capability has two benchmarks, this aggregation is equivalent to
the unweighted mean over the six development scores.
Table~\ref{tab:supp-candidate-backbones} compares every candidate on the same
current development partitions. TSVM provides the highest six-benchmark average and is selected as
$\boldsymbol{\theta}^{(1)}$. We then identify the weakest-preserved capability
as $w=\arg\max_{k\in\mathcal{K}}D_k(\boldsymbol{\theta}^{(1)})$, which selects
GUI interaction. Subsequent stages therefore use the TSVM recipient and GUI
expert evidence. The common evaluator assigns zero to TIES-Merging on the two
search partitions because its checkpoint yields no valid extracted answers.

\begin{table}[t]
\centering
\small
\setlength{\tabcolsep}{5pt}
\renewcommand{\arraystretch}{1.08}
\begin{tabular*}{\columnwidth}{@{\extracolsep{\fill}}lrrrr@{}}
\toprule
\textbf{Development set} & \textbf{WA} & \textbf{TA} & \textbf{TIES} &
\cellcolor{tablemethodblue}\textbf{TSVM} \\
\midrule
MMSearch20       & 45.0 & \textbf{50.0} & 0.0  & \cellcolor{tablemethodblue}\textbf{50.0} \\
FactualVQA120    & \textbf{42.5} & 37.5 & 0.0  & \cellcolor{tablemethodblue}40.0 \\
AndroidWorld24 & 20.8 & 16.7 & 25.0 & \cellcolor{tablemethodblue}\textbf{37.5} \\
OSWorld74        & \textbf{18.9} & 13.5 & 13.5 & \cellcolor{tablemethodblue}17.6 \\
V*Bench40        & 77.5 & 85.0 & 5.0  & \cellcolor{tablemethodblue}\textbf{87.5} \\
HR-Bench 8K80    & \textbf{72.5} & 68.8 & 65.0 & \cellcolor{tablemethodblue}71.3 \\
\midrule
\textbf{Average} & 46.2 & 45.2 & 18.1 & \cellcolor{tablemethodblue}\textbf{50.6} \\
\bottomrule
\end{tabular*}
\caption{Performance (\%) of all Stage-1 candidate backbones on the six fixed
development partitions. The average is the unweighted mean of the six scores;
WA and TA denote Weight Averaging and Task Arithmetic.}
\label{tab:supp-candidate-backbones}
\end{table}

\subsection{Implementation Hyperparameters}
\label{sec:supp-hyperparameters}

Table~\ref{tab:supp-hyperparameters} consolidates the concrete settings used
to construct the reported AgentPatch checkpoint. Ratios in Stage~3 are applied
per behavior group and per layer. The protection budget counts group-balanced
Guardian calibration examples, not protected neurons; the latter are obtained
by taking the union of the per-group selections.

\begin{table}[t]
\centering
\scriptsize
\setlength{\tabcolsep}{3.5pt}
\renewcommand{\arraystretch}{1.04}
\begin{tabular}{clp{0.57\columnwidth}}
\toprule
\textbf{Stage} & \textbf{Component} & \textbf{Setting} \\
\midrule
\rowcolor{tablegroupgray}
1 & Candidate operators & Weight Averaging, Task Arithmetic, TIES-Merging, and TSVM \\
1 & Task Arithmetic scale & $\lambda_{\mathrm{TA}}=0.3$ \\
\rowcolor{tablegroupgray}
1 & Parameter scope & Visual merger and language model \\
1 & Selected recipient & $\boldsymbol{\theta}^{(1)}$: TSVM \\
\midrule
\rowcolor{tablemethodblue}
2 & Weak-task expert & $\boldsymbol{\theta}_w$: GUI-Owl-7B \\
2 & Activity threshold & $\epsilon=10^{-5}$ \\
\rowcolor{tablemethodblue}
2 & Residual coefficient & $\beta=0.15$ \\
\midrule
3 & Calibration evidence & AndroidWorld24 and OSWorld74 \\
\rowcolor{dgcneutral}
3 & Repair groups & 4 AndroidWorld + 12 OSWorld \\
3 & Activation statistic & Absolute mean \\
\rowcolor{dgcneutral}
3 & Selection score & Output contribution \\
3 & Repair ratio $q_r$ & $0.0075$/AndroidWorld group; $0.0025$/OSWorld group \\
\rowcolor{dgcneutral}
3 & Guardian ratio $q_{\mathrm{guard}}$ & $0.01$ per protection group \\
3 & Guardian budget $B_g$ & 512 group-balanced calibration examples \\
\rowcolor{dgcneutral}
3 & Interpolation & $\alpha=0.02$ \\
3 & Layer scope & All 28 decoder FFN layers; no filter \\
\rowcolor{dgcneutral}
3 & Write unit & Whole FFN neuron (gate/up rows and down column; bias if present) \\
\rowcolor{dgcneutral}
3 & DGC analysis model & \texttt{gpt-5.5}; temperature $0$ \\
3 & DGC context & Full trajectories; no step/example truncation \\
\rowcolor{dgcneutral}
3 & DGC output & Strict JSON schema; 32,768-token generation limit \\
\bottomrule
\end{tabular}
\caption{Complete implementation settings for the reported AgentPatch model.}
\label{tab:supp-hyperparameters}
\end{table}

\section{DGC Workflow and Behavior Evidence Construction}
\label{sec:supp-dgc}

The Diagnoser--Guardian--Compiler (DGC) workflow translates execution-level
evidence into auditable behavior spans for the final repair stage. DGC
determines \emph{which behaviors} should be restored or protected, while the
behavior-conditioned activation procedure determines \emph{where} they are
represented. Table~\ref{tab:supp-dgc-contract} summarizes the evidence contract
that separates the three agents.

\subsection{Trajectory Evidence Pools}
\label{sec:supp-dgc-evidence}

For each calibration task $i$, we record its benchmark and task identifiers,
instruction, environment observations, serialized action sequence, and
evaluator outcome. The repaired recipient $\boldsymbol{\theta}^{(2)}$ and the
weak-task expert $\boldsymbol{\theta}_w$ are executed on the same task, forming
a paired record
\begin{equation}
    e_i=
    \left(
    b_i,u_i,\tau_i^{(2)},\tau_i^{w},
    y_i^{(2)},y_i^{w}
    \right),
\end{equation}
where $b_i$ and $u_i$ identify the benchmark and task, $\tau_i^{(2)}$ and
$\tau_i^{w}$ are the two trajectories, and $y_i^{(2)}$ and $y_i^{w}$ are their
evaluator outcomes. The two evidence pools are deliberately separated:

\begin{center}
\setlength{\fboxsep}{5pt}
\fcolorbox{dgcrepair}{dgcrepair}{\parbox{0.405\columnwidth}{
\centering\textbf{Repair Evidence Pool}\\[-1pt]
\scriptsize Paired same-task trajectories\\
$\boldsymbol{\theta}^{(2)}$: verified underperformance\\
$\boldsymbol{\theta}_{w}$: verified corrective behavior\\
\textit{Output: expert behavior spans}}}
\hfill
\fcolorbox{dgcprotect}{dgcprotect}{\parbox{0.405\columnwidth}{
\centering\textbf{Protection Evidence Pool}\\[-1pt]
\scriptsize Recipient-success bank\\
Correct weak-task behaviors\\
Complementary capabilities to preserve\\
\textit{Output: recipient behavior spans}}}
\end{center}

Repair evidence consists exclusively of paired records admitted to the fixed
strict-gap pool; support records are not promoted into repair targets. Complete
trajectories provide the environment state and execution context surrounding
each action.
Protection evidence instead comes from successful recipient behaviors and
complementary capabilities to preserve. All calibration tasks are disjoint from
the test partitions.

\subsection{Diagnoser: Mining Behavior-Critical Gaps}
\label{sec:supp-diagnoser}

The \emph{Diagnoser} aligns the recipient and expert trajectories and locates
a behaviorally meaningful divergence associated with the expert's
advantage. It jointly considers the task instruction, observations, action
arguments, evaluator feedback, and the actions following the divergence. The
result is expressed as a reusable workflow behavior rather than as a
task-specific command string. In the GUI-targeted instantiation reported here,
examples include establishing the correct input focus, invoking the required
interaction type, navigating the appropriate menu hierarchy, and confirming a
state-changing operation.

Each diagnosed group records its behavioral intent, target environment,
failure mode, matcher hints, and exact source records. The benchmark, task
identifier, and supporting action step make every repair behavior traceable to
an observed gap. Related gaps may share a behavior group, but all strict source
records are retained. Only expert-supported spans receive the \texttt{boost}
polarity; incorrect recipient actions assist diagnosis but are not compiled
into a negative update branch.

\subsection{Guardian: Constructing Protection Evidence}
\label{sec:supp-guardian}

The \emph{Guardian} identifies behaviors whose associated parameters should be
excluded from repair. It operates on the recipient-success bank and groups
evidence by the capability being preserved. In the GUI-targeted instantiation
reported here, protection evidence includes already-correct interaction
workflows and valid action formats, together with complementary capabilities
such as search-query formulation, evidence-grounded answers, fine-grained
visual judgment, high-resolution reasoning, and general tool-use schemas.
Failure-only records may reveal a potential interference risk, but they cannot
by themselves establish a protection group.

Every protection group retains typed provenance linking it to recipient- or
shared-success examples. Guardian proposes no weight changes and never converts
protected examples into repair targets. Its selectors instead define the
protected neuron union, so editing is constrained by evidence of functionality
already present in the recipient rather than by a generic distance heuristic.

\subsection{Compiler: Executable Span Selectors and Auditing}
\label{sec:supp-compiler}

The \emph{Compiler} converts the Diagnoser and Guardian outputs into two
canonical collections: repair groups and protection groups. A compiled group
contains a name, behavior tokens, executable matchers, source-task provenance,
and evidence references. Matchers use literal, regular-expression, parsed
action-field, or within-action sequence rules. Because activation calibration
serializes one action record at a time, a multi-step workflow description is
retained as group-level intent, while every selected span must be grounded in
an action that is actually present in the corresponding trajectory.

For a serialized action sequence $x_i$ and a matched character interval
$[a_{ir},b_{ir})$ from group $r$, tokenizer offset mappings convert the match
into behavior-token positions
\begin{equation}
    \mathcal{T}_{i,r}
    =
    \left\{
    t\;\middle|\;
    \operatorname{offset}(x_{i,t})
    \cap [a_{ir},b_{ir})\neq\emptyset
    \right\}.
    \label{eq:supp-span-compilation}
\end{equation}
The resulting repair positions select expert tokens, whereas protection
positions select recipient tokens. The Compiler narrows overly broad selectors
and records unresolved repair--protection conflicts; final protection is
enforced at the neuron level by Guardian exclusion. It also enforces two
fail-closed provenance checks: the union of repair-group sources must cover all
supplied strict gaps, and every protection group must be backed by
recipient-success evidence.

Table~\ref{tab:supp-dgc-trace} traces an AndroidWorld Markor calibration
example from a recipient--expert gap to an executable repair selector paired
with independent recipient-success protection evidence.

\begin{table}[H]
\centering
\scriptsize
\setlength{\tabcolsep}{3.5pt}
\renewcommand{\arraystretch}{1.1}
\begin{tabular}{p{0.22\columnwidth}p{0.69\columnwidth}}
\toprule
\textbf{Stage} & \textbf{Audited artifact} \\
\midrule
Observed gap & On \texttt{MarkorCreateNoteFromClipboard}, the recipient misses part of the focus--paste--save workflow while the expert succeeds. \\
\rowcolor{dgcrepair}
Diagnoser & Emits repair group \path{markor_filename_focus_paste_save} with expert-supported action steps. \\
\rowcolor{dgcprotect}
Guardian & Draws successful Markor-edit and valid mobile-action evidence from the independent recipient-success bank. \\
\rowcolor{dgcneutral}
Compiler & Compiles Diagnoser repair evidence and Guardian protection evidence into executable token selectors; repair spans overlapping protected behavior are narrowed or excluded. \\
\bottomrule
\end{tabular}
\caption{An audited DGC compilation trace. The repair and protection branches
originate from different trajectory records and meet only at Compiler overlap
resolution.}
\label{tab:supp-dgc-trace}
\end{table}

\begin{table*}[t]
\centering
\small
\setlength{\tabcolsep}{4pt}
\renewcommand{\arraystretch}{1.12}
\begin{tabular}{p{0.12\textwidth}p{0.24\textwidth}p{0.27\textwidth}p{0.29\textwidth}}
\toprule
\textbf{Agent} & \textbf{Evidence source} & \textbf{Compiled output} & \textbf{Constraint} \\
\midrule
\rowcolor{dgcrepair}
Diagnoser & Paired recipient--expert gaps & Repair behavior groups grounded in expert action spans & Only expert-supported behaviors are emitted; incorrect recipient actions do not form a negative update branch. \\
\rowcolor{dgcprotect}
Guardian & Recipient-success and shared-success evidence & Protection behavior groups grounded in recipient action spans & Failure-only records cannot establish protection; protected functionality must already be present. \\
\rowcolor{dgcneutral}
Compiler & Behavior groups and serialized trajectories & Executable token-span selectors with task/step provenance & Protection takes precedence over repair, and incomplete provenance fails closed. \\
\bottomrule
\end{tabular}
\caption{Evidence contract of the DGC workflow. Each agent has a distinct
evidence source, compiled artifact, and safety constraint.}
\label{tab:supp-dgc-contract}
\end{table*}

\subsection{DGC Prompt Templates}
\label{sec:supp-dgc-prompts}

The DGC agents use structured prompt templates with strict JSON outputs. The
templates below provide the parameterized instruction and input--output
framework used by DGC. Run-specific payloads---including trajectories, screenshots, and
behavior profiles---are replaced with descriptive placeholders in angle
brackets.

At the framework level, DGC accepts trajectories from any identified weak
capability. Since GUI interaction is identified as the weakest-preserved
capability in our experiments, the instantiated payload schemas below use
AndroidWorld and OSWorld trajectories as repair evidence, while search and
visual behaviors provide complementary protection evidence.

\begin{table*}[t]
\centering
\small
\setlength{\tabcolsep}{4pt}
\renewcommand{\arraystretch}{1.12}
\begin{tabular}{p{0.12\textwidth}p{0.24\textwidth}p{0.27\textwidth}p{0.29\textwidth}}
\toprule
\textbf{Agent} & \textbf{Prompt goal} & \textbf{Required input fields} & \textbf{Output schema anchor} \\
\midrule
\rowcolor{dgcrepair}
Diagnoser & Mine expert-directed behavior gaps without creating a suppression branch. &
\texttt{strict\_gap\_records}; full trajectory context; strict-gap coverage audit. &
\texttt{behavior\_gap\_groups}: intent, current failure, expert repair direction, matchers, source records, evidence refs. \\
\rowcolor{dgcprotect}
Guardian & Protect recipient-owned behavior using successful evidence rather than failure-only heuristics. &
Search/FVQA, visual, and GUI example sets separated into success, shared-success, current-only-success, and failure records. &
\texttt{protected\_behavior\_groups}: protected capability, tokens, matchers, source tasks, risk if modified, evidence refs. \\
\rowcolor{dgcneutral}
Compiler & Convert Diagnoser and Guardian groups into executable span selectors with auditable provenance. &
Agent outputs, matcher hints, full serialized trajectories, required strict-gap coverage list, and protection evidence. &
\texttt{repair\_token\_groups}, \texttt{protected\_token\_groups}, conflict rules, calibration policy, next repair spec. \\
\bottomrule
\end{tabular}
\caption{Prompt I/O schema summary for the reported DGC instantiation. The
templates expose the same information boundaries as the implementation: repair
is expert-directed, protection is recipient-success-grounded, and compilation
is deterministic and auditable.}
\label{tab:supp-dgc-prompt-contract}
\end{table*}

\noindent\textbf{Diagnoser prompt template.}
The Diagnoser prompt asks the agent to read paired recipient--expert weak-task
trajectories and identify behavior-critical gaps where the expert provides a
clear corrective behavior. The prompt explicitly disables negative
suppression: every editable target must move the recipient toward the expert on
expert-supported spans.

\begin{promptbox}{Prompt 1. Diagnoser behavior-gap template.}{dgcrepairdark}{dgcrepair}
\textbf{[SYSTEM PROMPT]}\par
You are the Behavior-Critical Gap Diagnoser for expert-directed weak-task
repair.
Read the supplied paired trajectories and identify behavior-critical gaps for
which the current merged model underperforms while the weak-task expert provides
a concrete corrective behavior. Use only the
supplied evidence. Compare the task instruction, observations, actions,
evaluator feedback, and subsequent environment states; do not infer an
unsupported success or failure.

Use only \texttt{strict\_gap\_records}. Every output group must cite at least
one supplied strict-gap record, and the union of
\texttt{source\_records} over all groups must cover every supplied strict gap.
When \texttt{context\_policy.mode} is \texttt{full}, inspect the complete
trajectory and retain all distinct provenance. Support records must not be
promoted into repair groups.

Describe reusable workflow behaviors rather than memorizing task titles.
Matchers must select the expert-side corrective action spans and must be
executable as literal, regular-expression, JSON-field, or within-action
sequence matchers. Prefer selectors that generalize across similar actions,
while retaining exact source records and evidence references. Emit no negative
suppression or protection groups. Every repair group must target the LLM and
set \texttt{target\_modules} to exactly \texttt{["llm"]}. Return one valid
JSON object and no surrounding Markdown.

\textbf{[USER PROMPT]}\par
\begin{scriptsize}
\begin{verbatim}
{
  "payload_version": "<PAYLOAD_VERSION>",
  "context_policy": {
    "mode": "full",
    "record_omission_allowed": false,
    "action_window": "complete_trajectory"
  },
  "selection_policy": {
    "primary_evidence":
      "current_bad_expert_better",
    "support_anchors": "disabled",
    "required_source_type": "strict_gap"
  },
  "strict_gap_records":
    <PAIRED_RECIPIENT_EXPERT_TRAJECTORIES>,
  "record_counts": <STRICT_GAP_COUNT_AUDIT>,
  "multimodal_context":
    <OPTIONAL_SCREENSHOT_EVIDENCE>
}
\end{verbatim}
\end{scriptsize}

\textbf{[OUTPUT: STRICT JSON]}\par
\begin{scriptsize}
\begin{verbatim}
{
  "agent_role": "behavior_gap_miner",
  "behavior_gap_groups": [{
    "name": "<GROUP_NAME>",
    "intent": "<REPAIR_INTENT>",
    "current_failure": "<OBSERVED_GAP>",
    "expert_repair_direction":
      "<EXPERT_SUPPORTED_BEHAVIOR>",
    "tokens": ["<TOKEN_OR_PHRASE>"],
    "matchers": [{
      "type": "literal|regex|json_field|sequence",
      "pattern": "<EXECUTABLE_PATTERN>",
      "field": "<OPTIONAL_JSON_FIELD>",
      "value": "<OPTIONAL_FIELD_VALUE>",
      "case_sensitive": false
    }],
    "target_benchmark": ["android_world|osworld"],
    "target_failure_modes": ["<FAILURE_MODE>"],
    "failure_type":
      "perception|cross_modal_alignment|reasoning|"
      "tool_action|mixed",
    "target_modules": ["llm"],
    "source_type": "strict_gap",
    "source_records": [{
      "benchmark": "android_world|osworld",
      "task_id": "<TASK_ID>"
    }],
    "evidence_refs": ["<TASK_AND_STEP_REFERENCE>"],
    "priority": <INTEGER>,
    "confidence": <VALUE_IN_[0,1]>,
    "rationale": "<EVIDENCE_GROUNDED_REASON>"
  }],
  "notes": ["<OPTIONAL_NOTE>"]
}
\end{verbatim}
\end{scriptsize}
\end{promptbox}

\noindent\textbf{Guardian prompt template.}
The Guardian prompt is deliberately grounded in behavior that the recipient
already performs correctly. This prevents the protection branch from becoming
a generic list of desirable capabilities and keeps protected regions tied to
observed recipient competence.

\begin{promptbox}{Prompt 2. Guardian protection template.}{dgcprotectdark}{dgcprotect}
\textbf{[SYSTEM PROMPT]}\par
You are the Guard Behavior Miner for protected repair. Identify recipient-owned
behaviors whose associated FFN neurons should be protected while repairing the
weak capability. Read cross-task evidence from search and FactualVQA,
fine-grained visual reasoning, and GUI interaction.

Build protection groups primarily from \texttt{success\_examples},
\texttt{current\_success\_examples}, \texttt{shared\_success\_examples}, and
\texttt{current\_only\_success\_examples}. Failure or expert-better examples
may explain risk, but they cannot by themselves establish that the recipient
owns a behavior. Never describe a failure-only pattern as a protected success.
When full context is supplied, inspect every typed success set and preserve
distinct evidence provenance.

Protect search-query formulation, evidence-grounded answers, numerical
precision, visual attribute and spatial reasoning, and valid GUI action
schemas when supported by successful recipient evidence. Every protection
group must retain exact evidence references and executable matchers that can
select recipient-side calibration spans. Do not propose repair, boost, or
suppression targets. Set \texttt{target\_modules} to \texttt{["llm"]} for the
reported decoder-FFN patch. Return one valid JSON object and no surrounding
Markdown.

\textbf{[USER PROMPT]}\par
\begin{scriptsize}
\begin{verbatim}
{
  "payload_version": "<PAYLOAD_VERSION>",
  "context_policy": {
    "mode": "full",
    "example_omission_allowed": false
  },
  "current_model": "<RECIPIENT_MODEL>",
  "task_profiles": {
    "search": <SEARCH_BEHAVIOR_PROFILES>,
    "vision": <VISUAL_BEHAVIOR_PROFILES>,
    "gui": <GUI_BEHAVIOR_PROFILES>
  },
  "multimodal_context":
    <OPTIONAL_SCREENSHOT_METADATA>,
  "gui_pairwise_evidence":
    <OPTIONAL_PAIRED_GUI_CONTEXT>
}
\end{verbatim}
\end{scriptsize}

Each task profile contains its metrics and typed \texttt{example\_sets}; the
actual success, shared-success, current-only-success, and failure examples are
inserted at the placeholders above.

\textbf{[OUTPUT: STRICT JSON]}\par
\begin{scriptsize}
\begin{verbatim}
{
  "agent_role": "guard_behavior_miner",
  "protected_behavior_groups": [{
    "name": "<GROUP_NAME>",
    "intent": "<PROTECTION_INTENT>",
    "tokens": ["<TOKEN_OR_PHRASE>"],
    "matchers": [{
      "type": "literal|regex|json_field|sequence",
      "pattern": "<EXECUTABLE_PATTERN>",
      "field": "<OPTIONAL_JSON_FIELD>",
      "value": "<OPTIONAL_FIELD_VALUE>",
      "case_sensitive": false
    }],
    "source_tasks": ["<SOURCE_BENCHMARK>"],
    "protected_capability": "<CAPABILITY>",
    "target_modules": ["llm"],
    "risk_if_modified": "<INTERFERENCE_RISK>",
    "evidence_refs": ["<SUCCESS_EVIDENCE_REF>"],
    "priority": <INTEGER>,
    "confidence": <VALUE_IN_[0,1]>
  }],
  "notes": ["<OPTIONAL_NOTE>"]
}
\end{verbatim}
\end{scriptsize}
\end{promptbox}

\noindent\textbf{Compiler prompt template.}
The Compiler prompt converts the two mined evidence sets into the exact
selector format consumed by activation calibration. Its central role is to
preserve provenance while preventing broad or conflicting repair selectors
from overriding protected behavior.

\begin{promptbox}{Prompt 3. Compiler selector template.}{dgcneutraldark}{dgcneutral}
\textbf{[SYSTEM PROMPT]}\par
You are the Selector Compiler for expert-directed behavior-gap repair. Combine
the Diagnoser behavior-gap groups and Guardian protection groups into the
executable selector specification consumed by activation calibration. Preserve
valid matcher fields and convert matcher hints into literal, regular-expression,
JSON-field, or within-action sequence matchers.

Compile Diagnoser groups as repair behavior spans with
\texttt{polarity="boost"} and \texttt{target\_span\_role="positive"}; compile
Guardian groups as protection behavior spans with
\texttt{polarity="protect"} and \texttt{target\_span\_role="protected"}. This
workflow has no negative suppression branch. All repair and protection groups
target the LLM in the reported decoder-FFN patch.

Protection takes precedence over repair. If a repair selector overlaps a
protected behavior, narrow the repair selector or record the conflict for hard
Guardian exclusion. Avoid broad selectors that match generic action tokens.
Because each calibration text serializes one action, do not emit multi-action
workflow descriptions as sequence matchers; every repair group must retain a
matcher that directly selects its source action JSON.

Preserve \texttt{source\_type}, \texttt{source\_records}, matchers, and
\texttt{evidence\_refs}. The union of compiled repair sources must cover every
required strict-gap record. The canonical groups are
the top-level \texttt{repair\_token\_groups} and
\texttt{protected\_token\_groups}; keep the auxiliary selector lists in
\texttt{next\_repair\_spec} empty to prevent duplication. Return raw JSON only.

\textbf{[USER PROMPT]}\par
\begin{scriptsize}
\begin{verbatim}
{
  "context_policy": {
    "mode": "full",
    "group_field_omission_allowed": false,
    "provenance_omission_allowed": false
  },
  "agent_b": {
    "agent_role": "behavior_gap_miner",
    "behavior_gap_groups": <DIAGNOSER_OUTPUT>,
    "notes": <DIAGNOSER_NOTES>
  },
  "agent_g": {
    "agent_role": "guard_behavior_miner",
    "protected_behavior_groups": <GUARDIAN_OUTPUT>,
    "notes": <GUARDIAN_NOTES>
  },
  "required_source_records":
    <STRICT_GAP_COVERAGE_LIST>,
  "serialized_trajectories":
    <CALIBRATION_TRAJECTORIES>
}
\end{verbatim}
\end{scriptsize}

\textbf{[OUTPUT: STRICT JSON]}\par
\begin{scriptsize}
\begin{verbatim}
{
  "agent_role": "selector_compiler",
  "repair_token_groups": [{
    "name": "<GROUP_NAME>",
    "tokens": ["<TOKEN_OR_PHRASE>"],
    "polarity": "boost",
    "target_span_role": "positive",
    "matchers": [<EXECUTABLE_MATCHER_OBJECTS>],
    "target_benchmark": ["android_world|osworld"],
    "target_failure_modes": ["<FAILURE_MODE>"],
    "failure_type": "<FAILURE_TYPE>",
    "target_modules": ["llm"],
    "source_type": "strict_gap",
    "source_records": [<SOURCE_RECORDS>],
    "priority": <INTEGER>,
    "confidence": <VALUE_IN_[0,1]>,
    "evidence_refs": ["<TASK_AND_STEP_REFERENCE>"]
  }],
  "protected_token_groups": [{
    "name": "<GROUP_NAME>",
    "tokens": ["<TOKEN_OR_PHRASE>"],
    "polarity": "protect",
    "target_span_role": "protected",
    "matchers": [<EXECUTABLE_MATCHER_OBJECTS>],
    "source_tasks": ["<SOURCE_BENCHMARK>"],
    "protected_capability": "<CAPABILITY>",
    "target_modules": ["llm"],
    "reason": "<INTERFERENCE_RISK>",
    "evidence_refs": ["<SUCCESS_EVIDENCE_REF>"]
  }],
  "conflict_resolution": {
    "rules": ["<PROTECTION_PRECEDENCE_RULE>"],
    "resolved_conflicts": ["<CONFLICT_AUDIT>"]
  },
  "calibration_policy": {
    "positive_examples": ["<REPAIR_EXAMPLE_REF>"],
    "negative_or_protected_examples":
      ["<PROTECTION_EXAMPLE_REF>"],
    "selection_rule": "<SELECTION_RULE>"
  },
  "next_repair_spec": {
    "selector_groups": [],
    "protected_selector_groups": [],
    "suggested_modules": ["llm"],
    "notes": ["<OPTIONAL_NOTE>"]
  }
}
\end{verbatim}
\end{scriptsize}
\end{promptbox}

This prompt design makes the analysis agents inspect rich execution evidence
while keeping the compiled artifact small and machine-checkable. The
workflow additionally applies deterministic validation to verify strict-gap
coverage, success-grounded Guardian provenance, and executable matcher fields.
The downstream patcher uses only the frozen selectors and trajectories; the
prompt templates are not invoked during model inference.

\subsection{Compiled Evidence Audit}
\label{sec:supp-dgc-audit}

After compilation, selectors and task/step provenance are frozen. The selector
audit in Table~\ref{tab:supp-dgc-audit} reports the compiled artifacts consumed
by teacher-forced activation collection and whole-neuron interpolation.

\begin{table}[t]
\centering
\small
\setlength{\tabcolsep}{5pt}
\renewcommand{\arraystretch}{1.08}
\begin{tabular}{lr}
\toprule
\textbf{Compiled selector artifact} & \textbf{Count} \\
\midrule
Repair groups after Compiler & 16 \\
Protection groups & 22 \\
Expert repair spans & 272 \\
Recipient protection spans & 291 \\
Inference-time DGC calls & 0 \\
\bottomrule
\end{tabular}
\caption{Compiled selector audit used to construct the final static checkpoint.}
\label{tab:supp-dgc-audit}
\end{table}

These artifacts are used only during offline construction of the static
AgentPatch checkpoint and introduce no analysis-agent dependency at inference
time.

\section{Additional Analyses}
\label{sec:supp-additional-analyses}

\subsection{Layer-Wise Distribution of Repair Regions}
\label{sec:supp-repair-regions}

\begin{figure}[t]
\centering
\includegraphics[width=\columnwidth]{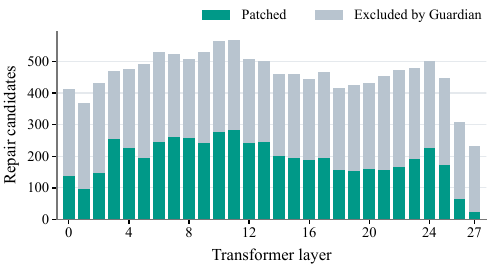}
\caption{Layer-wise distribution of repair candidates before and
after Guardian exclusion.}
\label{fig:supp-repair-regions}
\end{figure}

Figure~\ref{fig:supp-repair-regions} visualizes where the behavior-critical
patch acts across the 28 decoder layers. Repair candidates occur throughout the
network rather than concentrating in a narrow block of late layers, indicating
that the localization procedure selects repair candidates across the network.
Guardian
exclusion removes a substantial subset in every layer while retaining a
smaller behavior-directed patch. In total, the 12,887 pre-Guardian candidates
are reduced to 5,352 patched neurons. This distribution illustrates that the
Guardian does not simply protect a small set of exceptional layers; it
constrains editing wherever repair evidence overlaps behavior regions
associated with capabilities to preserve.

\subsection{Decoupling Single- and Cross-Environment Repair Evidence}
\label{sec:supp-domain-evidence}

\begin{figure}[H]
\centering
\includegraphics[width=\columnwidth]{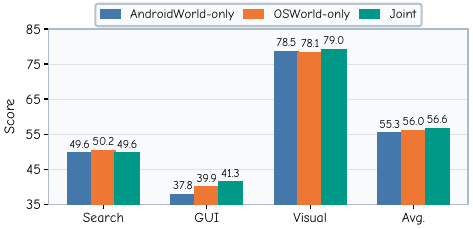}
\caption{Effect of using AndroidWorld-only, OSWorld-only, and joint
repair evidence under the same final budget of 5,352 repair neurons.}
\label{fig:supp-domain-evidence}
\end{figure}

Figure~\ref{fig:supp-domain-evidence} isolates the contribution of combining
behavior evidence from AndroidWorld and OSWorld. AndroidWorld-only and
OSWorld-only repair are budget-matched to Full AgentPatch, so their comparison
does not benefit from editing different numbers of neurons. The joint variant
achieves the strongest GUI, Visual, and overall results, reaching 41.3, 79.0,
and 56.6, respectively. These results show that the two evidence sources
provide complementary behavioral coverage: jointly compiling their repair
groups yields a more balanced patch than specializing the same editing budget
to either environment alone.

\subsection{Qualitative Examples}
\label{sec:supp-qualitative-examples}

\begin{figure*}[p]
\centering
\includegraphics[width=0.97\textwidth]{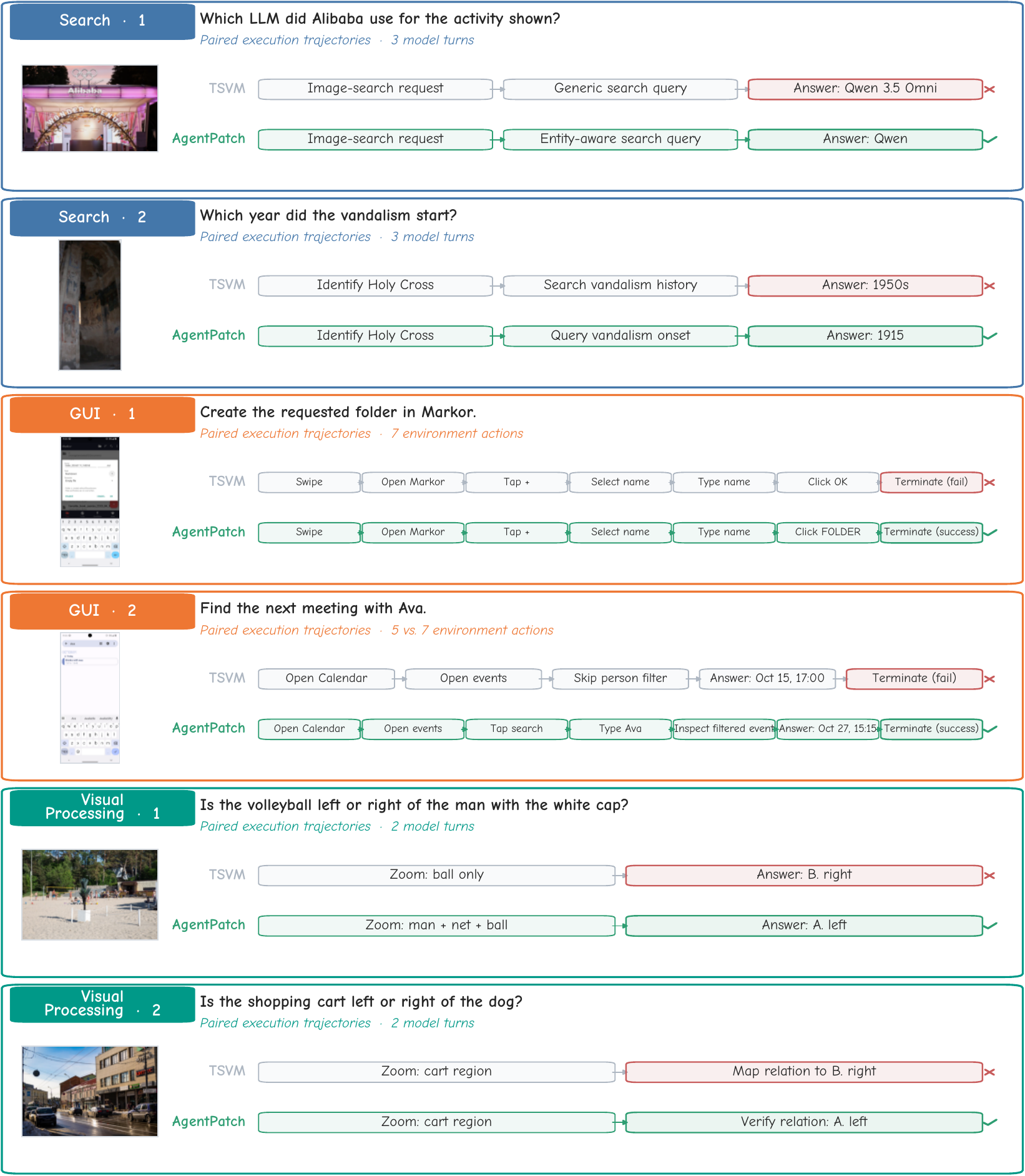}
\caption{Qualitative comparison between TSVM and AgentPatch across Search, GUI
interaction, and Visual Processing. Each task family contains two paired
evaluation examples. The boxes summarize recorded model, tool, or environment
actions, and the displayed trajectory lengths follow the corresponding
executions. Red and green endpoints denote unsuccessful and successful task
outcomes, respectively.}
\label{fig:supp-repair-examples}
\end{figure*}

Figure~\ref{fig:supp-repair-examples} presents paired execution examples from
the three agentic task families. Across the two search cases, AgentPatch forms
or interprets the decisive retrieval evidence more precisely, leading to the
correct entity or event-time answer. In GUI interaction, the repaired model
retains behavior-critical operations---selecting the folder action and applying
the required person filter---that determine whether the requested workflow is
completed. The visual-processing cases further show that preserving the
relevant visual context and correctly mapping the inspected spatial relation
can change the final outcome. Together, these examples illustrate that the
repair is reflected in complete agentic trajectories rather than only in their
final answers.

\section{Limitations}
\label{sec:supp-limitations}

AgentPatch currently operates on experts derived from a shared base model with
aligned parameter structures, consistent with standard model-merging settings.
Constructing the behavior-critical patch requires offline calibration
trajectories and analysis-agent calls, introducing a one-time preprocessing
cost; however, the resulting model remains a single static checkpoint with no
additional inference overhead. Our experiments focus on merging three expert
models, and future work could evaluate the scalability of the same workflow
under merging settings with different numbers of experts.

\FloatBarrier

\end{document}